\documentclass[letterpaper,10pt,conference]{ieeeconf}  

\IEEEoverridecommandlockouts
\usepackage{amsmath} 
\usepackage{amssymb}
\usepackage{amsfonts}

\usepackage{multirow}
\usepackage{multicol}
\usepackage{booktabs}
\usepackage{graphicx}

\usepackage{balance}
\usepackage{cite}
\usepackage{cleveref}
\Crefname{figure}{Fig.}{Figs.} 
\usepackage{url}

\usepackage{amsthm}
\newtheoremstyle{remboldstyle}
  {}{}{\itshape}{}{\bfseries}{.}{.5em}{{\thmname{#1 }}{\thmnumber{#2}}{\thmnote{ (#3)}}}
\theoremstyle{remboldstyle}

\newtheorem{theorem}{Theorem}[section]
\newtheorem{definition}{Definition}[section]

\usepackage{xparse}

\NewDocumentCommand\bbm{}{ \begin{bmatrix} }
\NewDocumentCommand\ebm{}{ \end{bmatrix} }
\NewDocumentCommand\Vector{m}{ \boldsymbol{\mathbf{#1}} }
\NewDocumentCommand\Matrix{m}{ \boldsymbol{\mathbf{#1}} }

\NewDocumentCommand\ArgMax{m}{ \operatorname*{argmax}_{#1} }

\NewDocumentCommand\Real{}{ \mathbb{R} }

\NewDocumentCommand\LieGroupSE{m}{ \mathrm{SE}(#1) }

\NewDocumentCommand\CoordinateFrame{m}{ \underrightarrow{\Matrix{\mathcal{F}}}_{#1} }

\newcommand{\tinyframe}[1]{{\scriptscriptstyle #1}}

\usepackage{xcolor}

\title{\LARGE \bf Observability-Aware Trajectory Optimization: \\Theory, Viability, and State of the Art}

\author{Christopher Grebe, Emmett Wise, and Jonathan Kelly$^{\dag}$ 
\thanks{All authors are with the Space \& Terrestrial Autonomous Robotics Systems (STARS) Laboratory at the University of Toronto Institute for Aerospace Studies and Robotics Institute, Toronto, Ontario, Canada {\tt <firstname>.<lastname>@robotics.utias.utoronto.ca}.}
	\thanks{$^\dag$Jonathan Kelly is a Vector Institute Faculty Affiliate. This research was supported in part by the Canada Research Chairs program.}}

\begin{document} 


\maketitle 
\thispagestyle{empty}
\pagestyle{empty}

\begin{abstract}
Ideally, robots should move in ways that maximize the knowledge gained about the state of both their internal system and the external operating environment. 
Trajectory design is a challenging problem that has been investigated from a variety of perspectives, ranging from information-theoretic analyses to leaning-based approaches.
Recently, observability-based metrics have been proposed to find trajectories that enable rapid and accurate state and parameter estimation.
The viability and efficacy of these methods is not yet well understood in the literature. 
In this paper, we compare two state-of-the-art methods for observability-aware trajectory optimization and seek to add important theoretical clarifications and valuable discussion about their overall effectiveness.
For evaluation, we examine the representative task of sensor-to-sensor extrinsic self-calibration using a realistic physics simulator.
We also study the sensitivity of these algorithms to changes in the information content of the exteroceptive sensor measurements.
\end{abstract}

\section{Introduction}

Robots generally use a combination of exteroceptive and interoceptive sensing to collect task-relevant information about the external environmental and their internal state. 
If the data collected are insufficient, a robot may perform unreliably or, in the worst case, become a safety hazard.
For example, many self-calibration algorithms will return faulty results if the data do not meet certain excitation conditions \cite{kelly_visual-inertial_2011}.
As a result, we require methods to maximize the amount of relevant information (about the states and parameters of interest) that is provided by the data collected. 

A variety of methods have been developed to increase the information content of (i.e., the knowledge provided by) a robot trajectory.
Factors that influence the information obtained include the environment, system dynamics, process and measurement noise, and the expressiveness of the trajectory parameterization, among others. 
%
Recently, the notion of using nonlinear observability as a criterion for trajectory optimization has been proposed in the literature.
Typically, nonlinear observability is treated as a binary test that determines if the measurements and inputs of an ideal (noise-free) system are sufficient to recover a unique solution for the states and parameters.
Krener and Ide \cite{krener_measures_2009} were the first to suggest the `degree of observability' as a metric for trajectory optimization. 
Their method relies on costly simulation and numerical integration, however.
Recent approaches by Preiss et al.\ \cite{preiss_simultaneous_2018} and Frey et al.\ \cite{frey_online_2020} avoid these computational difficulties, to some extent, by developing alternative observability-based metrics.
Preiss et al.\ \cite{preiss_simultaneous_2018} introduce a novel method to maximize a function related to the empirical local observability Gramian \cite{krener_measures_2009}; Frey et al.\ \cite{frey_online_2020} take a different approach in the stochastic setting by introducing an interval filtering model and metric (described in \Cref{sec:stochastic}).
Importantly, optimization metrics based on nonlinear observability naturally account for the nonlinear form of the system and measurement models, potentially leading to better results than competing techniques.

\begin{figure}[t]
\vspace{2mm}
\centering
\includegraphics[width=\columnwidth]{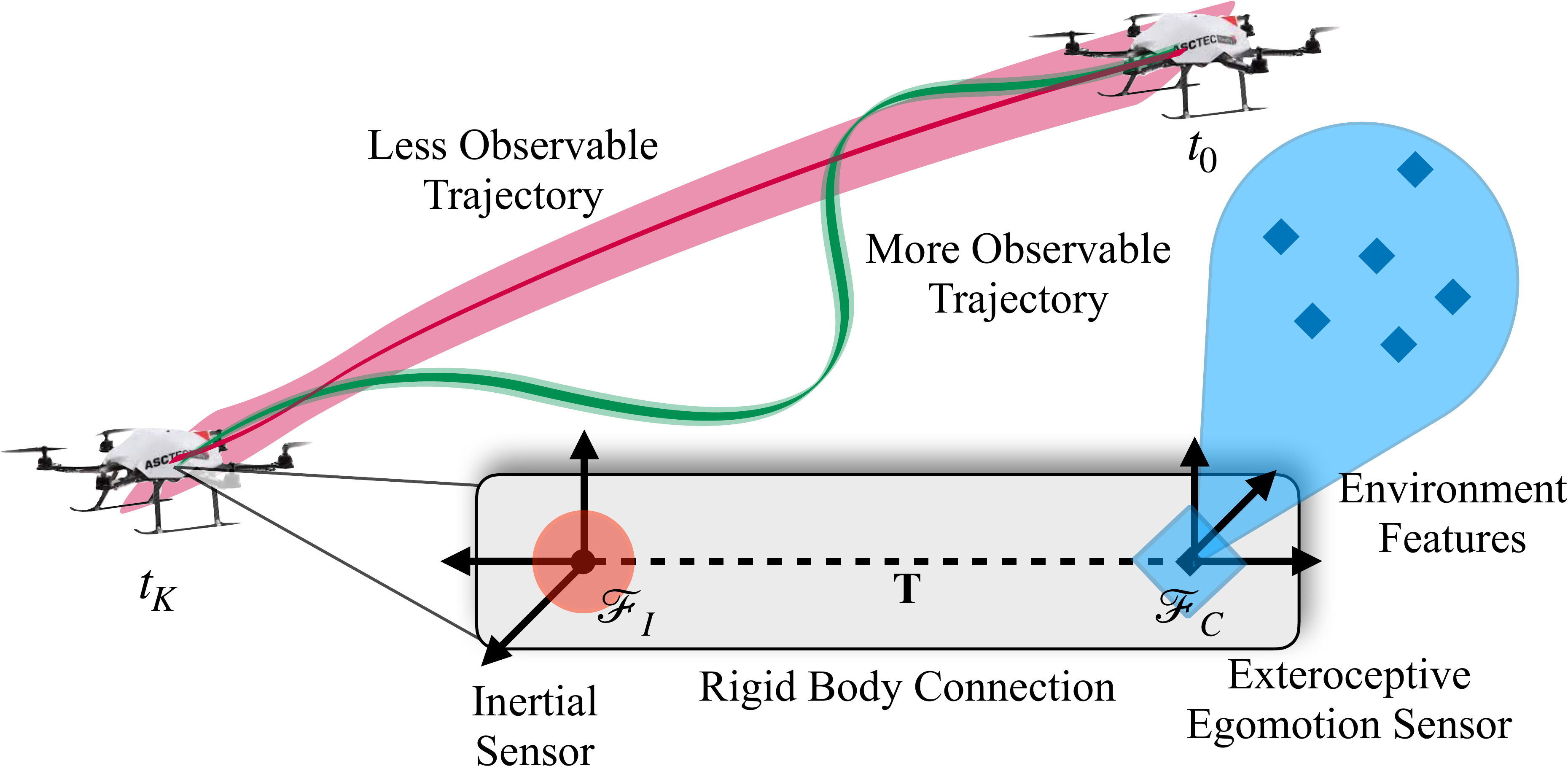}
\caption{Highly-observable trajectories can improve estimation accuracy during active self-calibration. As an example, we may wish to estimate the transformation $\Matrix{T}_{\tinyframe{IC}}\! \in \LieGroupSE{3}$ from the egomotion sensor (camera) frame $\CoordinateFrame{\tinyframe{C}}$ to the inertial sensor frame $\CoordinateFrame{\tinyframe{I}}$ while simultaneously tracking the pose of the robot platform.}
\label{fig:main_fig}
\vspace{-3.5mm}
\end{figure}

To the best of the authors' knowledge, there is no existing comprehensive analysis of the performance of techniques of the type introduced by Preiss et al.\ \cite{preiss_simultaneous_2018} and Frey et al.\ \cite{frey_online_2020}.
Bohm et al.\ \cite{bohm_observabilty-aware_2020} determine the performance of the method in \cite{preiss_simultaneous_2018} for a real robot, but do not review the work in  \cite{frey_online_2020}.
%
In this paper, we compare the methods of Preiss et al.\ \cite{preiss_simultaneous_2018} and Frey et al.\ \cite{frey_online_2020} on the representative task of self-calibration of the rigid-body transform between an inertial measurement unit (IMU) and a stereo camera (or other sensor that is able to provide egomotion estimates).
This paper contributes the following: 
\begin{itemize}
\item a comprehensive review of important works related to observability-aware trajectory optimization;
\item a comparison of two state-of-the-art observability-aware trajectory optimization algorithms, made under controlled conditions in a realistic physics simulation environment;
\item a sensitivity analysis of the algorithms to variations in the `quality' of exteroceptive measurements; and
\item a discussion of the viability of the use of observability to inform trajectory optimization. 
\end{itemize}

The paper is structured as follows.
We provide a survey of related literature in \Cref{sec:related_work}.
In \Cref{sec:problem_formulation}, we review our mathematical notation and present the generic trajectory optimization problem. 
\Cref{sec:observability} describes in detail the concept of observability as it relates to this work, while \Cref{sec:methodology} considers the methodologies of Preiss et al.\ \cite{preiss_simultaneous_2018} and of Frey et al.\ \cite{frey_online_2020}. 
Finally, we present and discuss our results in \Cref{sec:exp-results} and offer some conclusions in \Cref{sec:conclusion}. 

\section{Related Work}
\label{sec:related_work}

The problem of trajectory optimization has received substantial attention in the literature and a wide variety of optimization techniques exist.
One the most common methods for trajectory optimization in robotics is to maximize a norm of the Fisher information matrix (FIM) \cite{schneider_visual-inertial_2017,albee_combining_2019,wilson_trajectory_2014,zhang_fisher_2020}.
The aim of such approaches is to find a trajectory that increases the amount of information available from exteroceptive sensor measurements about the relevant system states and parameters.
However, this knowledge also depends on the system dynamics and the nonlinear form of the  measurement model equations, which are important additional considerations not fully addressed by FIM-type metrics.

In contrast to the optimization of metrics related to the FIM, the optimization of an observability criterion \cite{krener_measures_2009,preiss_simultaneous_2018,frey_online_2020} accounts for nonlinear system dynamics and measurements.
One of the first observability-related metrics introduced in the literature was the integration-based metric defined by Krener and Ide in \cite{krener_measures_2009}.
This metric is based on the \emph{empirical local observability Gramian}, which quantifies the change in the measured outputs when the initial system states are perturbed while the inputs are held constant. 
A drawback of the approach in \cite{krener_measures_2009} and similar techniques is that each evaluation of the objective function requires the integration of two \cite{krener_measures_2009} or more \cite{zeng_observability_2018} ordinary differential equations (ODEs), which is computationally very challenging.
	
Powel and Morgansen \cite{powel_empirical_2020} apply stochastic differential equations, instead of ODEs, to account for process noise (i.e., model uncertainty) when determining observability. 
In \cite{powel_empirical_2020}, the empirical local observability Gramian framework is extended to the setting of \emph{stochastic observability}: there is still a rank condition, but it is now defined in terms of the expectation over a distribution.
The authors describe how noise on the control inputs can potentially improve nonlinear observability---in some cases, noise may excite a system in ways that the control inputs cannot.
However, the method in \cite{powel_empirical_2020} is restricted to stochastic systems with closed-form solutions; in the absence of a closed-form solution,  computationally intensive and numerically unstable simulations are usually required \cite{preiss_simultaneous_2018}.

Without a computationally tractable and reliable method to maximize some measure of the observability of a trajectory, the majority of existing work has relied on other techniques: maximizing information gain, minimizing posterior uncertainty, or other ad hoc criteria.
For example, Maye et al.\ \cite{maye_online_2016} evaluate the information gained  from sensor measurements during trajectory execution by comparing the mutual information of the current trajectory segment with previously-stored trajectory segments.
Additionally, the algorithm in \cite{maye_online_2016} filters out trajectory segments for which the state is unobservable, based on the conditioning of the FIM.
Schneider et al.\ develop a similar approach in \cite{schneider_visual-inertial_2017}, utilizing differential entropy to quantify the information content of each trajectory segment.
Usayiwevu et al.\ \cite{usayiwevu_information_2020} minimize the posterior covariance of the extrinsic transform parameters for a lidar-IMU self-calibration task through informative path planning.
The environment is first explored and mapped while following a pre-defined trajectory.
This initial trajectory is then extended using sampling-based motion planning \cite{bry_rapidly-exploring_2011} to minimize the parameter uncertainty. 

\section{Problem Formulation}
\label{sec:problem_formulation}

In this section, we introduce the notation used throughout the paper, describe our representative calibration task, and review the general trajectory optimization problem.

\subsection{Notation} 
\label{subsec:notation}

Lowercase Latin and Greek letters (e.g., $a$ and $\alpha$) denote scalar variables, while uppercase letters are are reserved for sets.
Boldface lower and uppercase letters (e.g., $\Vector{x}$ and $\Matrix{\Theta}$) denote vectors and matrices, respectively. 
A three-dimensional reference frame is designated by $\CoordinateFrame{}$.
The translation vector from point $a$ (often a reference frame origin) to a point $b$, expressed in $\CoordinateFrame{a}$, is denoted by $\!\Vector{p}_{ab}$. 
When there is no risk of ambiguity, we denote a vector quantity, such as linear velocity or acceleration, or angular velocity, defined in a specific reference frame, with a single subscript.
For example, the angular velocity of frame $\CoordinateFrame{a}$, expressed in $\CoordinateFrame{a}$, is denoted as $\Vector{\omega}_a$.
The bold lowercase symbol $\Vector{q}_{ab}$ is the unit quaternion that defines the orientation of $\CoordinateFrame{b}$ with respect to $\CoordinateFrame{a}$.
We use the notation $\Matrix{R}(\Vector{q}_{ab})$ for the operator that converts the unit quaternion $\Vector{q}_{ab}$ to the respective $\mathrm{SO}(3)$ rotation matrix.
The notation $\mathcal{N}(\Vector{\mu}, \Matrix{\Sigma})$ denotes a multivariate Gaussian distribution with a mean vector $\Vector{\mu}$ and covariance matrix $\Matrix{\Sigma}$.

\subsection{Sensor-to-Sensor Extrinsic Self-Calibration} 
\label{subsec:ext-cal}

The task we examine in \Cref{sec:exp-results} is extrinsic self-calibration between an inertial measurement unit (IMU) and a stereo camera (or a similar sensor that is able to estimate egomotion). 
Three reference frames are involved: the \textit{world} frame, $\CoordinateFrame{\tinyframe{W}}$, a frame anchored to the Earth, the \textit{IMU} frame, $\CoordinateFrame{\tinyframe{I}}$, aligned with the centre of the sensor, and the \textit{camera} frame, $\CoordinateFrame{\tinyframe{C}}$, fixed at the middle of the stereo camera. This configuration is depicted in \Cref{fig:main_fig}.
The state vector is
\begin{equation}
	\Vector{x}(t) = 
	\bbm 
	\Vector{p}_{\tinyframe{WI}}(t)^T &\!\!
	\Vector{q}_{\tinyframe{WI}}(t)^T &\!\!
	\Vector{v}_{\tinyframe{W}}(t)^{T}\!\! & 
	\Vector{b}_g(t)^T\!\! & 
	\Vector{b}_a(t)^T
	\ebm^T,
\end{equation}
where $\Vector{p}_{\tinyframe{WI}}(t)$ is the vehicle position, $\Vector{q}_{\tinyframe{WI}}(t)$ is the unit quaternion that defines the vehicle attitude, $\Vector{v}_{\tinyframe{W}}(t)$ is the vehicle velocity, and $\Vector{b}_g(t)$ and $\Vector{b}_a(t)$ are the inertial sensor gyroscope and accelerometer biases, respectively.
The extrinsic calibration parameters are 
\begin{equation}
	\Theta = 
	\bbm 
	\Vector{p}_{\tinyframe{IC}}^T &\!
	\Vector{q}_{\tinyframe{IC}}^T 
	\ebm^T,
\end{equation}
where $\Vector{p}_{\tinyframe{IC}}\!$ is the position of the camera relative to the IMU and $\Vector{q}_{\tinyframe{IC}}\!$ is the unit quaternion that defines the camera orientation relative to the IMU frame.
The state evolves in time according to
\begin{align}
\dot{\Vector{p}}_{\tinyframe{WI}}(t) & = 
\Vector{v}_{\tinyframe{W}}(t),  &
\dot{\Vector{q}}_{\tinyframe{WI}}(t) & = 
\frac{1}{2}\Omega\big(\Vector{\omega}_\tinyframe{I}(t)\big)\,\Vector{q}_{\tinyframe{WI}}(t),\\
\dot{\Vector{v}}_{\tinyframe{W}}(t) & = 
\Vector{a}_{\tinyframe{W}}(t),\\
\dot{\Vector{b}}_g(t) & = \Vector{n}_{gw}(t), &
\dot{\Vector{b}}_a(t) & = \Vector{n}_{aw}(t),\\
\dot{\Vector{p}}_{\tinyframe{IC}} & = \Vector{0}_{3 \times 1}, &
\dot{\Vector{q}}_{\tinyframe{IC}} & = \Vector{0}_{4 \times 1},
\end{align}
where $\Omega(\cdot)$ is the quaternion kinematic matrix \cite{kelly_visual-inertial_2011}, $\Vector{\omega}_\tinyframe{I}$ is the angular velocity of the IMU expressed in the IMU frame, and $\Vector{a}_{\tinyframe{W}}$ is the acceleration of the IMU expressed in the world frame. The gyroscope and accelerometer biases are assumed to be uncorrelated  white Gaussian random walk processes defined by vectors $\Vector{n}_{gw} \sim \mathcal{N}(0, \Matrix{\Sigma}_{gw})$ and $\Vector{n}_{aw} \sim \mathcal{N}(0, \Matrix{\Sigma}_{aw})$, respectively. 
The measured IMU angular velocity and acceleration (kinematic inputs) are
\begin{align}
	\Vector{\omega}_{m}(t) & = \Vector{\omega}_{\tinyframe{I}}(t) + 
	\Vector{b}_{g}(t)+\Vector{n}_{g}(t),\\
	\Vector{a}_{m}(t) & = \Vector{R}^T(\Vector{q}_{\tinyframe{WI}}(t))\big(\Vector{a}_{\tinyframe{W}}(t)-\Vector{g}\big)+\Vector{b}_{a}(t) + \Vector{n}_{a}(t),
\end{align}
respectively, where $\Vector{g}$ is the gravity vector. The vectors $\Vector{n}_{g} \sim\mathcal{N}(0, \Matrix{\Sigma}_g)$ and $\Vector{n}_{a} \sim \mathcal{N}(0, \Matrix{\Sigma}_a)$ are uncorrelated gyroscope and accelerometer white Gaussian noise terms, respectively.
The stereo camera (or other exteroceptive sensor) is able to measure its position and orientation,
\begin{equation}
	\Vector{h}_1(t) = \Vector{p}_{\tinyframe{WI}}(t) + \Vector{R}\big(\Vector{q}_{\tinyframe{WI}}(t)\big)\,\Vector{p}_{\tinyframe{IC}},
\end{equation}
and
\begin{equation}
	\Vector{h}_2(t) = \Vector{q}_{\tinyframe{WI}}(t) \otimes \Vector{q}_{\tinyframe{IC}},
\end{equation}
relative to the world frame, where $\otimes$ denotes quaternion multiplication.
We note that these are \emph{indirect} measurements that can be derived from camera observations of a sufficient number of landmarks in the environment. For brevity, we omit discussion of the imaging process (which depends on the camera geometry) and the associated image noise.

\subsection{Trajectory Optimization}
\label{sec:opt-max}

For completeness, we define the the full trajectory optimization problem in the context of maximum a posterior estimation as described by Maye et al. \cite{maye_self-supervised_2013}.
For the system defined in \Cref{subsec:ext-cal}, a discretized trajectory $\Xi(X, \Theta, U, Z)$ is given by 
the states $X = \{\Vector{x}_{k}\, |\, k = 0,\dots,K\}$, 
the calibration parameters $\Theta$,
the inputs $U = \{\Vector{u}_k\, |\, k = 1,\dots,K\}$, and
the measurements $Z = \{\Vector{z}_k\, |\, k = 1,\dots,K\}$.
We seek to determine the posterior distribution $p(X, \Theta\, |\, U, Z)$. 
This joint posterior can be factored as
\begin{equation}
p(X, \Theta\, |\, U, Z) \propto \prod_{k=1}^K p(\Vector{x}_k\, |\, \Vector{x}_{k-1}, \Vector{u}_k) \prod_{k=1}^K p(\Vector{z}_{k}\, |\, \Vector{x}_k, {\Theta}).
\end{equation}
Conventionally, the full joint posterior is assumed to be a normal distribution with mean $\Vector{\mu}$ and covariance $\Vector{{\Sigma}}$.
The maximum a posteriori estimate of the mean vector is
\begin{equation}
\label{eq:map-opt}
	\Vector{\mu}_{X, \Theta} = \ArgMax{X, \Theta} p(X, \Theta\, |\, U, Z).
\end{equation}
The trajectory optimization problem is to find an optimized trajectory that maximizes some metric that depends on one or more components of the vector $\Vector{\mu}_{X, \Theta}$ (such as the calibration parameters) or the covariance matrix $\Matrix{\Sigma}_{X, \Theta}$.

\begin{figure*}[t]
\centering
\includegraphics[trim={0 7cm 0 0},clip,width=\textwidth]{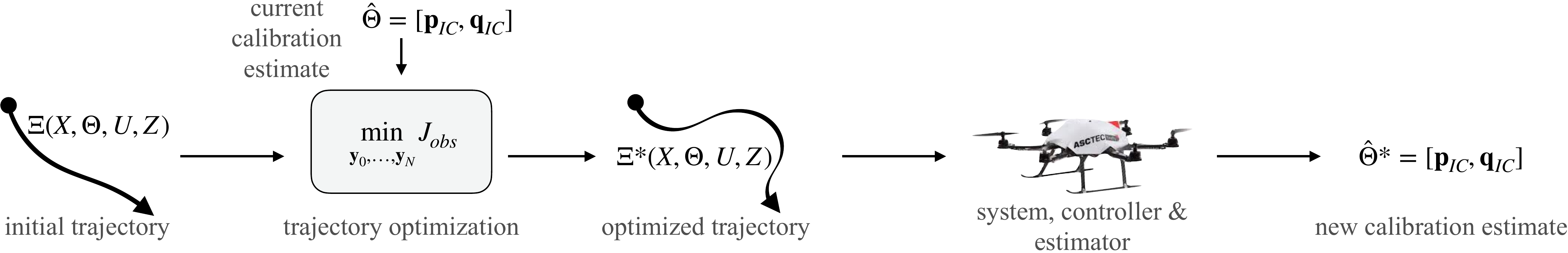}
\caption{System optimization pipeline. An initial trajectory is optimized based on a chosen observability metric. The optimized trajectory is then executed by the controller in a realistic physics simulation environment to determine overall performance. We use a quadrotor drone in our simulations.
}
\vspace{-2mm}
\label{fig:system}
\end{figure*}

\section{Observability} 
\label{sec:observability}

In this section, we present the theoretical foundations necessary to understand the use of observability as a trajectory optimization metric.
The section begins with a review of nonlinear observability, followed by a discussion of the relationship between the observability matrix and trajectory optimization from two perspectives.
First, the observability-trajectory relationship is discussed through the lens of observability analyses in the presence of noisy measurements and controls.
Second, observability and the system trajectory are related by a limit on the knowledge that the trajectory and associated measurements provide about states or parameters being estimated.
Martinelli \cite[pg.\ 68]{martinelli2020observability} defines weak observability as follows.
\begin{definition}[Weak Observability]
	An input-output system is weakly observable at $\Vector{x}(t_0)$ if there exists an open set $B$ of $\Vector{x}(t)$ such that, by knowing that $\Vector{x}(t_0) \in B$, there exists at least one choice of inputs $\Vector{u}(t)$ such that $\Vector{x}(t_0)$ can be obtained from the knowledge of the output $\Vector{h}(t)$ and the inputs $\Vector{u}(t)$ on the time interval $\mathcal{I}$.
\end{definition}

The analysis that determines the weak observability of a system begins with a Taylor series expansion of the control input and measurement functions, starting at the time $t_0$ \cite{martinelli2020observability},
\begin{equation} \label{eq:meas_taylor}
	\Vector{h}(t)=\left. \sum_{i=0}^{\infty} \frac{d^{i} \Vector{h}(t)}{d t^{i}} \right|_{t=t_0} \frac{(t-t_0)^{i}}{i !}
\end{equation}
and
\begin{equation}
	\Vector{u}(t)=\left. \sum_{i=0}^{\infty} \frac{d^{i} \Vector{u}(t)}{d t^{i}} \right|_{t=t_0}\frac{(t-t_0)^{i}}{i !}.
\end{equation}
Importantly, this formulation implies that knowledge of the measurements and control inputs over an interval is equivalent to knowledge of all of the measurement and input derivatives at $\Vector{x}(t_0)$.
Using the Taylor expansion, the system of equations related to the unknown state $\Vector{x}(t_0) = \Vector{x}_{0}$ is
\begin{equation} \label{eq:sys-eqns}
	\bbm \Vector{h}(t_0) \\ \Vector{\dot h}(t_0) \\ \Vector{\ddot h}(t_0) \\ \vdots \ebm = 
	\bbm  L^0(\Vector{x}_{0},\Vector{u}(t_0)) \\  L^1(\Vector{x}_{0},\Vector{u}(t_0)) \\ L^2(\Vector{x}_{0}, \Vector{u}(t_0)) \\ \vdots \ebm,
\end{equation}
where $i$th Lie derivative $L^i$ is defined as
\begin{equation}
\label{eqn:lie_deriv}
L^i(\Vector{x},\Vector{u}) = \frac{\partial\,L^{i - 1}\left(\Vector{x}, \Vector{u}\right)}{\partial\,\Vector{x}}
\Vector{f}(\Vector{x},\Vector{u}),
\end{equation}
and where we drop the time dependence for clarity. The $i$th Lie derivative is the gradient of the previous Lie derivative with respect to the state, in the direction of the vector field determined by the system dynamics $\Vector{f}(\Vector{x},\Vector{u})$. We take $L^0 = \Vector{h}(\Vector{x}_0)$ to be the measurement function itself.
Note that this approach inherently accounts for nonlinearities in both the dynamics and the measurement models.
The Jacobian of \eqref{eq:sys-eqns} is the well-known nonlinear observability matrix $\Matrix{\mathcal{O}}_{0}$ introduced by Hermann and Krener in \cite{hermann_nonlinear_1977},
\begin{equation} \label{eq:obs}
\Matrix{\mathcal{O}}_0 \triangleq
\bbm \nabla \Vector{h}(t_0) \\ \nabla \Vector{\dot h}(t_0) \\ \nabla \Vector{\ddot h}(t_0) \\ \vdots \ebm = 
	  \bbm  \nabla L^0(\Vector{x}_0,\Vector{u}(t_0)) \\  \nabla L^1(\Vector{x}_0,\Vector{u}(t_0)) \\  \nabla L^2(\Vector{x}_0,\Vector{u}(t_0)) \\ \vdots \ebm.
\end{equation}
Determining \eqref{eq:obs} is often computationally intractable because the matrix contains an infinite number of Lie derivatives (rows).
In practice, a finite order of Lie derivative can be selected to make the computation feasible.

To recover the initial state $\Vector{x}_0$, the matrix \eqref{eq:obs} must invertible.
However, real-world controls and measurements contain noise.
Noise impacts the conditioning of \eqref{eq:obs}, or the degree of observability.
Since measurement noise, in particular, reduces the degree of observability, this noise reduces the ability to determine $\Vector{x}(t_0)$.
The only inputs available to the system to mitigate the effects of noise are the set of controls. 

From Martinelli \cite[pg.\ 71]{martinelli2020observability}, a knowledge bound for nonlinear observability is presented as follows. 

\begin{theorem}[Lie Derivative Knowledge Bound] 
\label{thm:knowledge}
The knowledge of $\Vector{x}(t_0) = \Vector{x}_0$ gathered from $\Vector{u}(t)$ and $\Vector{h}(t)$ is enclosed in the values of all the Lie derivatives of the system at $\Vector{x}_0$. In other words, knowledge of the values of all of the Lie derivatives of the system at $\Vector{x}(t_0)$ is an upper bound on the knowledge of $\Vector{x}_0$. 
\end{theorem}
By incorporating nonlinear observability in a trajectory optimization framework \cite{preiss_simultaneous_2018}\cite{frey_online_2020}, the solution of the trajectory optimization problem maximizes the knowledge about $\Vector{x}_0$ (or some subset of the state vector).

\section{Methodology} \label{sec:methodology}

The differential-geometric approach to nonlinear observability inherently accounts for nonlinear system dynamics.
Both Preiss et al.\ \cite{preiss_simultaneous_2018} and Frey et al.\ \cite{frey_online_2020} leverage higher-order approximations of the measurement model `Jacobian' through the use of Lie derivatives, 
\begin{equation} \label{eq:meas_approx}
	\nabla \Vector{h}_0(t) = \left. \sum_{i=0}^{\infty}  \nabla L^i(\Vector{x}(t),\Vector{u}(t)) \right |_{t = t_0} \frac{(t-t_0)^i}{i!}
\end{equation}
where a subscript is added (e.g, $\Vector{h}_0(t)$) to clarify the point at which the Lie derivatives are evaluated. 
In the subsections below, we describe and contrast the approaches in \cite{preiss_simultaneous_2018} and \cite{frey_online_2020}.
An overview of our full optimization pipeline is shown in \Cref{fig:system}.
We note that, despite measurement uncertainty being known to influence observability \cite{huang_observability_2020} and calibration accuracy \cite{huang_geometric_2018} for robotic systems, neither \cite{preiss_simultaneous_2018} and \cite{frey_online_2020} account for measurement uncertainty.

\subsection{Deterministic Method}
\label{sec:deterministic}

As discussed in \Cref{sec:related_work}, Preiss et al.\  \cite{preiss_simultaneous_2018} develop a deterministic trajectory optimization metric based upon an earlier, integration-based technique \cite{krener_measures_2009}.
A matrix, known as the \emph{expanded empirical local observability Gramian} (or E$^2$LOG) matrix, ${\Vector{\widetilde A}}_{n,H}$, is first computed for the time window $[\bar t_n, \bar t_{n+1} = \bar t_n + H]$ where $\bar t_n$ is the interval start time and $H$ is a positive multiple of the exteroceptive sensor measurement period $\Delta t$. 
The metric integrates the measurement model Jacobian defined by \eqref{eq:meas_approx}, 
\begin{equation}
	{\Vector{\widetilde A}}_{n,H} = \int_{0}^{H} \nabla \Vector{h}_n(t)^T\nabla \Vector{h}_n(t)\,dt.
\end{equation}
The matrix can be written equivalently as, 
\begin{equation} \label{eq:final_simp_e2log}
		{\Vector{\widetilde A}}_{n,H}=  
		\Vector{\mathcal{O}}_{n}^T \Vector{W}_1
 \Vector{\mathcal{O}}_{n},
\end{equation} 
where
\begin{equation}  
	\Vector{W}_1 = \bbm \frac{H^{i+j+1}}{(i+j+1)i!j!} \Vector{I} & \hdots \\ \vdots & \ddots \ebm.
\end{equation}
The matrix $\Vector{\mathcal{O}}_{n}$ is defined by \eqref{eq:obs} and computed at time $\bar{t}_n$. 
In the weighting matrix, $\Matrix{W}_1$, $i$ and $j$ are the orders of the Lie derivative contained in the observability matrices on the left and the right in \eqref{eq:final_simp_e2log}, respectively.
The complete metric is determined by summing over $N$ intervals along a trajectory, each of length $H$ and with its own unique start time $\bar{t}_{{n}}$, 
\begin{equation}
	\Vector{A}_\Xi = \sum_{n = 0}^{N} \Vector{\widetilde A}_{n,H},
\end{equation}
resulting in the E$^2$LOG matrix, $\Vector{A}_\Xi$, for the trajectory. 
The actual metric is one of: the smallest singular value of $\Vector{A}_\Xi$, its trace, or the condition number of the matrix.

\subsection{Stochastic Method} 
\label{sec:stochastic}

Frey et al.\ \cite{frey_online_2020} devise an alternative trajectory optimization approach that explicitly considers the stochastic nature of robotic systems.
The method of \cite{frey_online_2020} is based on information filtering, that is, a variant of Bayesian filtering in information form. 
The starting point is the discrete-time system error state model, 
\begin{align}
	\Vector{e}_{k+1} & = \Matrix{\tilde F}_k \Vector{e}_k + \Matrix{\tilde G}_k\Vector{w}_k,\\
	\Vector{\tilde z}_k & = \Matrix{H}_k\Vector{e}_k + \Vector{n}_k,
\end{align}
where $\Vector{e}_k$ is the error state vector, $\Vector{\tilde z}_k$ is the measurement vector, $\Vector{w}_k$ is the process noise vector, and $\Vector{n}_k$ is the measurement noise vector.
The matrix $\Matrix{F}_k$ is the motion model Jacobian, the matrix $\Matrix{H}_k$ is the measurement model Jacobian, and the matrix $\Matrix{G}_c$ is the (continuous-time) process noise Jacobian.
Finally, $\Matrix{\tilde F}_k = \Matrix{I}+\Delta t\, \Matrix{F}_k$ is the error state transition matrix and $\Matrix{\tilde G}_k = \sqrt{\Delta t}\,\Matrix{G}_c$ is the matrix that maps the process noise from continuous time to discrete time.

As shown previously, the measurement Jacobian can be written using \eqref{eq:meas_approx}. 
However, the Lie derivative formulation of \eqref{eq:meas_approx} does not account for process noise. 
Frey et al.\ solve this problem by incorporating the process model uncertainty before the Lie derivative approximation is applied,
\begin{equation}
 	\Vector{e}_{n}^{+} = \Matrix{E}_n \left[\begin{array}{l}
\mathbf{e}_{n} \\ 
\mathbf{w}_{n}
\end{array}\right],
 \end{equation}
 \begin{equation}
 	\Matrix{E}_n = {\left[\begin{array}{ll}
1\! & \left.\sqrt{H} \Matrix{\Phi}_{0}^{K} \Vector{G}_{k}\right]
\end{array}\right.},
 \end{equation}
where 
$\Matrix{\Phi}_{0}^{K} \approx \Matrix{F}_K\Matrix{F}_{K-1} \dots \Matrix{F}_0$. 
The matrix $\Matrix{E}_n$ injects process noise `immediately' at the start of the time window $[\bar t_n, \bar t_{n+1} = \bar t_n + H]$ (as opposed to incrementally at each time step within the window).
Throughout the window, exteroceptive measurements are received at times $t_k$, where $t_k$ is determined relative to the interval start time $\bar t_n$. 
With use of  \eqref{eq:meas_approx}, the error state measurement $\Vector{\tilde z}_k$ is then
\begin{equation}
	\Vector{\tilde z}_k = \Matrix{L}_n \bbm \Vector{e}_n \\ \Vector{w}_n \ebm + \Vector{n}_k,
\end{equation} 
\begin{equation}
\label{eq:temp_var}
	\Matrix{L}_n = 
	\nabla \Vector{h}_n(t) \Matrix{E}_n,
\end{equation}
where $\nabla \Vector{h}_n(t)$ is defined by \Cref{eq:meas_approx} and where we omit the dependence of $\Matrix{L}_n$ on the variable $t$ for brevity.
For the interval $[\bar t_n, \bar t_{n+1}]$, the information matrix over $\Vector{e}_n$ and $\Vector{w}_n$ is 
\begin{equation}
	\Vector{\widetilde B}_{n,H} = \Matrix{L}_n^T\Matrix{L}_{n}  = \Vector{E}_n^T \Vector{\mathcal{O}}_n^T \Vector{W}_2\Vector{\mathcal{O}}_n\Vector{E}_n,
\end{equation}
where
\begin{equation}
	\Vector{W}_2 = \bbm \frac{t_{k=1}^{i+j} + t_{k=2}^{i+j} + \hdots }{i!j!} \Vector{I} & \hdots \\ \vdots & \ddots \ebm. 
\end{equation} 
To propagate to the next time window, Frey et al.\ use $\mathbf{e}_{n+1} \approx \Matrix{\Phi}_{0}^{K} \mathbf{e}_{n}^{+}$.
Thus, 
\begin{equation}
	\Vector{B}_{n+1} = (\Matrix{\Phi}^K_0)^{-T} \left[ \Vector{B}_n + \Vector{\widetilde B}_{n,H} \right](\Matrix{\Phi}^K_0)^{-1},
\end{equation}
where joint information matrix at the start of an interval with an initial covariance $\Matrix{P}_0$ is
\begin{equation}
	\Vector{B}_n = \bbm \Vector{P}_0^{-1} & \Vector{0} \\ \Vector{0} & \Vector{I}\ebm.
\end{equation}
The above process is repeated for $N$ subsequent intervals until the final matrix $\Matrix{B}_\Xi$ is obtained.
Similarly to the E$^2$LOG criterion, the metric is then defined by the smallest singular value, trace, or condition number of $\Matrix{B}_\Xi$. 

\subsection{Trajectory Parameterization} 
\label{sec:traj_param}

In \Cref{sec:problem_formulation}, we defined the trajectory parameters for our calibration task as $\Vector{p}_{\tinyframe{WI}}(t) \in \Real^3$ and $\Vector{q}_{\tinyframe{WI}}(t) \in \mathrm{S}^3$.
A drawback of this parameterization in the optimization context is that it may result in the generation of dynamically infeasible trajectories.
To prevent this problem, in our experiments, we consider the flight of a quadrotor-type vehicle and recast the trajectory parameterization in a differentially flat form \cite{mellinger_minimum_2011}. 
Information on the conversion of the $\Real^3$ and $\mathrm{S}^3$ poses to the differentially flat form can be found in \cite{mellinger_minimum_2011}. 

We require a continuous-time representation of the trajectory also, for two reasons.
First, the trajectory must be sampled at arbitrary times to evaluate the optimization metrics.
Second, the conversion from differentially flat space to $\Real^3$ and $\mathrm{S}^3$ requires the instantaneous derivatives of the differentially flat parameters with respect to time. 
Due to the linearity of differentially flat dynamics, we use a uniform $\Real^4$-spline parameterization of the continuous differentially flat trajectory. 

A uniform $\Real^d$-spline represents the trajectory using a finite set of knots $\Vector{Y} = \{ \Vector{y}_i | i=0, \dots, N\}$, a spline order $k$, and a temporal spacing between knots $\Delta t_{knot}$.
Each knot is assigned a time $t_i = t_0 + i\Delta t_{knot}$.
When evaluating the spline value at time $t$, the time is normalized to $u = \frac{t - t_i}{\Delta t_{knot}}$, where $t_i \leq t < t_{i+1}$.
From Sommer et al.\ \cite{sommer_efficient_2020}, an $\Real^4$-spline can be represented using
\begin{equation} \label{eq:spline_def}
\Vector{y}(u) = 
\bbm \Vector{y}_i\! & \Vector{d}_1^i\! & \cdots\! & \Vector{d}_{k-1}^i \ebm \Matrix{\tilde{M}}^{(k)}\Vector{u},
\end{equation}
where $\Vector{y}_i$ is a knot, $\Vector{d}_j^i = \Vector{y}_{i + j} - \Vector{y}_{i + j - 1}$, $\Matrix{\tilde{M}}^{(k)}$ is a mixing matrix and $\Vector{u}^T = \bbm 1\! & u\! & \cdots\! & u^{k-1} \ebm$.

\subsection{Trajectory Optimization}
\label{subsec:trajectory-optimization}

The observability-aware optimization cost function $J_{obs}$ can be defined for the matrices $\Matrix{A}_\Xi$ or $\Matrix{B}_\Xi$ from \Cref{sec:deterministic} and \Cref{sec:stochastic}, respectively. 
In this work the matrix trace is applied, 
\begin{equation}
	J_{obs} = \textbf{tr}(\Matrix{A}_\Xi \Matrix{S} ),
\end{equation}
\begin{equation}
	J_{obs} = \textbf{tr}(\Matrix{B}_\Xi \Matrix{S}),
\end{equation}
where $\Matrix{S}$ is an additional selection matrix that determines which states or parameters the optimization is applied to.

Similar to \cite{zhou_raptor:_2021-1}, the convex hull property of splines is used to enforce four constraints in the optimization: a fixed starting position $\Vector{y}_{start}$, a fixed ending position $\Vector{y}_{end}$, a maximum velocity $\Vector{v}_{max}$, and a maximum acceleration $\Vector{a}_{max}$. 
The convex hull property ensures that the value of a spline at time $t$ must lie in the convex hull of the knots $t_i$ to $t_{i + k - 1}$.
This property is leveraged to ensure that the starting and ending spline values are within a slack parameter $\Vector{\epsilon}$ of their desired values.
Conveniently, the convex hull property also extends to the time derivatives of the spline.
As a result, the velocity knots and acceleration knots constrain the velocity and acceleration along the spline as well.
The full optimization problem is then defined as,
\begin{equation}
\label{eq:optimization-problem}
\begin{aligned}
\min_{\Vector{y}_0, ..., \Vector{y}_N} \quad & J_{obs} \\
\textrm{s.t.} \quad & |\dot{\Vector{y}}_i| \leq \Vector{v}_{max} \; \forall \; i \geq 1,\\
& |\ddot{\Vector{y}}_i| \leq \Vector{a}_{max} \; \forall \; i \geq 2,\\
& |\Vector{y}_j - \Vector{y}_{end}| \leq \Vector{\epsilon}_{} \; \text{for} \; j =N - k + 1, \dots, N,\\
& |\Vector{y}_j - \Vector{y}_{start}| \leq \Vector{\epsilon}_{} \; \text{for} \; j =0, \dots,k-1. 
\end{aligned}
\end{equation}

For our comparison in \Cref{subsec:results}, we implemented one additional optimization metric based on an acceleration term and similar to that found in Zhou et al. \cite{zhou_raptor:_2021-1}. The cost functional is
\begin{equation} 
\label{eq:accel_cost}
J_{accel} = \int_{0}^{t_N + \Delta t} \Vector{\ddot{y}}(t)^T\Matrix{W}_3\,\Vector{\ddot{y}}(t) dt, 
\end{equation}
where $\Matrix{W}_3$ is a user-specified weight matrix (which we simply set to the identity matrix herein).
Since acceleration and deceleration require energy, this metric provides an indication of the energy expended by the system while executing a specific trajectory.
Minimizing \eqref{eq:accel_cost} effectively minimizes the maximum acceleration (as well as the average acceleration) along the trajectory, and so we refer trajectories optimized in this way as MMA trajectories.

\section{Experiments} 
\label{sec:exp-results}

The deterministic and stochastic trajectory optimization methods described herein can be applied to many estimation problems.
We chose to compare these algorithms on the benchmark task of IMU-to-stereo camera extrinsic self-calibration.
This is a well-studied problem because accurate calibration is important for overall system performance.
We describe our simulation environment and setup in \Cref{subsec:setup}, present our experimental results in \Cref{subsec:results}, and discuss those results in \Cref{subsec:discussion}.

\subsection{Simulation Setup} 
\label{subsec:setup}

To determine the relative performance of the trajectory optimization methods, we performed two simulation experiments.
In the first experiment, we compared the accuracy of the calibration results obtained using observability-optimized trajectories against the results obtained using random and MMA trajectories.
Our second experiment analyzed the sensitivity of observability-optimized calibration to variations in the quality (i.e., information content) of the stereo camera measurements.

Experiments were carried out in the Gazebo simulator \cite{koenigDesignUseParadigms2004} with the RotorS MAV plugin \cite{furrer2016rotors}.
A modified version of the multi-state constraint Kalman filter (MSCKF), available in the OpenVINS \cite{geneva_openvins:_2020} software package, was used as our state estimator.
OpenVINS allows the number of visible camera landmarks to be easily adjusted, with new landmarks generated randomly within the camera field of view as needed.
Trajectory optimization was carried out using the nonlinear solver IFOPT \cite{winkler2018gait}.
To manipulate the trajectory splines, we used the \texttt{basalt-headers} library \cite{sommer_efficient_2020}.

For each experimental test, a random spline was generated between a fixed starting point and a randomly-chosen end point. 
The random spline included fifteen knots, spaced evenly in time at half-second intervals.
Starting with the random initial spline, we the applied the IFOPT solver with three different criteria: minimizing the maximum acceleration along the trajectory (MMA), optimizing the deterministic metric of Preiss et al., and optimizing the stochastic metric of Frey et al. 
The optimization time window $H$, defined in \Cref{sec:deterministic}, is the period over which the observability component of the trajectory optimization metrics are recalculated.
We used a time window of $H = 0.2$ s as a compromise between the $0.1$ s window suggested  by Preiss et al.~\cite{preiss_simultaneous_2018} and the $\leq 0.3$ s window used by Frey et al.~\cite{frey_online_2020}.

\subsection{Results} 
\label{subsec:results}

\begin{figure}[]
\centering  
\includegraphics[width=0.7\columnwidth]{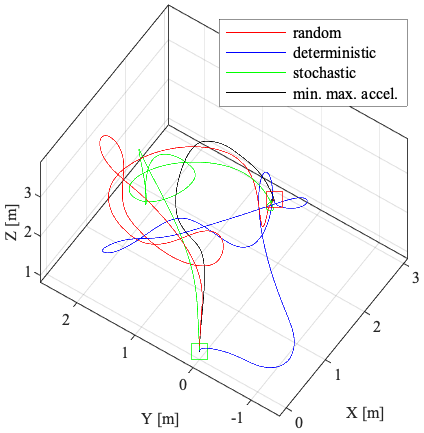}
\caption{Example of trajectories from one simulation trial: random, deterministic  \cite{preiss_simultaneous_2018}, stochastic \cite{frey_online_2020}, and minimized maximum acceleration. The deterministic and stochastic trajectories were optimized using the respective observability-aware algorithms for translation accuracy. The start and goal positions are indicated by green and red squares, respectively. 
}
\label{fig:traj_comp}
\vspace{-4mm}
\end{figure}

\begin{figure}[]
\centering  
\includegraphics[trim={0 0.5mm 0 0},clip,width=0.9\columnwidth]{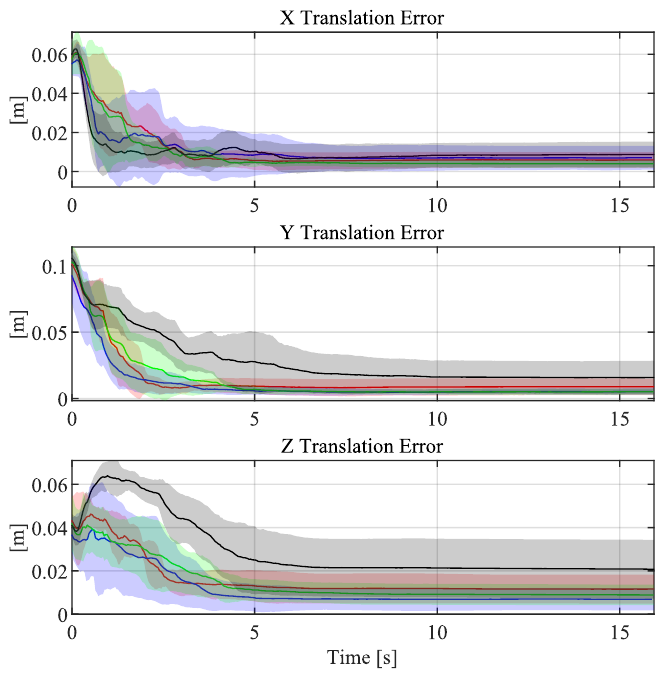}
\caption{Comparison of extrinsic calibration accuracy for random (red), deterministic  \cite{preiss_simultaneous_2018} (blue), stochastic \cite{frey_online_2020} (green), and minimized maximum acceleration (black) trajectories.
For each method, the mean performance across six trials is indicated by the solid line and the one-sigma standard deviation is indicated by the shaded area. 
}
\label{fig:comp_trans_all}
\vspace{-4mm}
\end{figure}

We compared the performance of the observability-aware trajectory optimization methods over six simulation trials.
We also considered MMA trajectories since this class of trajectories is used in many many robotics problems \cite{mellinger_minimum_2011,zucker_chomp:_2013}.
\Cref{fig:traj_comp} provides a visualization of the trajectories produced during one simulation trial. 
In \Cref{fig:comp_trans_all}, we present a comparison of the extrinsic translation calibration accuracy over all test cases with 400 landmarks visible per image frame.
The sensor-to-sensor translation parameters are the most difficult to calibrate in general \cite{kelly_visual-inertial_2011}.
For all trials we kept the random image plane measurement noise at a standard deviation of one pixel, in line with the noise value used in \cite{geneva_openvins:_2020}. 
Both the deterministic and stochastic observability-aware methods lead to significantly better calibration accuracy than the randomly-excited and MMA trajectories.
The benefit of the observability-aware algorithms is most apparent when considering the calibration accuracy of the z-axis offset.
This is the translation along the axis that lies parallel to the camera optical axis, which is known to be a difficult value to determine reliably \cite{kelly_visual-inertial_2011}.

\begin{figure}[]
\centering  
\includegraphics[trim={0 0.5mm 0 0},clip,width=0.9\columnwidth ]{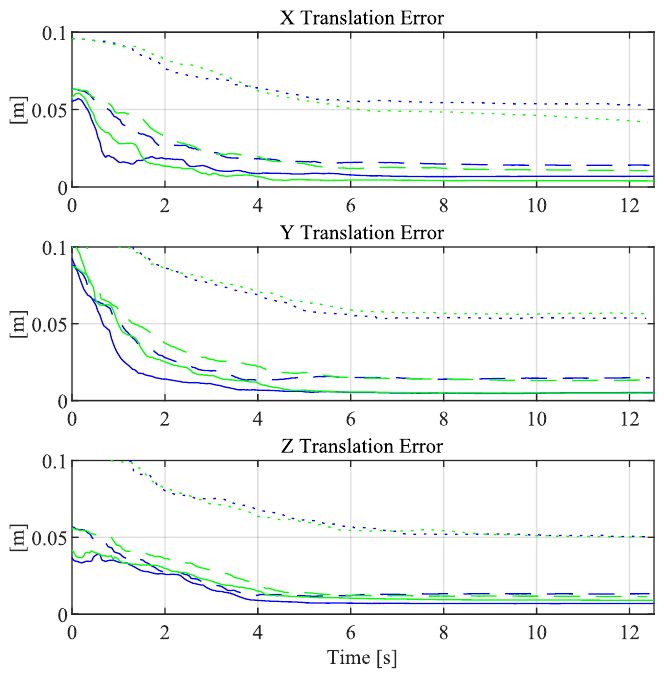}
\caption{Comparison of extrinsic calibration accuracy using the metric in \cite{preiss_simultaneous_2018} (blue) and in \cite{frey_online_2020} (green) for high (solid), medium (dashed), and low (dotted) exteroceptive measurement quality. We  optimized for extrinsic translation accuracy only.
For both methods and all levels of measurement quality, the mean performance across six trials is shown.
}
\label{fig:ext_trans_meas}
\vspace{-5mm}
\end{figure}

We also investigated the impact of exteroceptive measurement 
`quality' on the accuracy of the observability-aware algorithms.   
We defined quality in terms of the number of landmarks (and image features) visible in each image frame.
Dynamic and consistent adjustment of the number of visible landmarks was made possible by the OpenVINS software---in contrast, photorealistic simulators cannot guarantee repeatable and consistent exteroceptive measurements across different trajectories. 
In \Cref{fig:ext_trans_meas},  self-calibration performance for varying levels of exteroceptive measurement quality is compared.
Trials with 4, 40, and 400 landmarks per image frame were evaluated; these were designated as low, medium, and high quality respectively.
It can be seen that the most significant drop in calibration accuracy occurs when exteroceptive measurement quality is reduced from medium to low, as would be expected.

Lastly, we considered a measure of trajectory cost in our comparison.
We determined an `acceleration cost' through a forward finite difference approximation of the changing acceleration along each trajectory.
This cost was simply the sum of the squared acceleration values at all knots along the trajectory; we then normalized the cost values by diving by the average cost of the most expensive trajectories.
\Cref{tab:cost} presents the normalized cost, sum of squared estimation error, and cost-normalized error for each set of trajectories. 
The sum of squared estimation error is the squared error for the extrinsic translation parameters (only), summed over each trajectory.
The cost-normalized error is the product of the normalized cost and the sum of squared estimation error. 
This error value allows for a comparison of the estimation accuracy against the average cost of a trajectory.

\begin{table}[b]
\centering
\caption{Performance comparison for calibration of the extrinsic translation parameters with varying levels of exteroceptive measurement quality.}
\label{tab:cost}
\setlength{\tabcolsep}{5pt}
\renewcommand{\arraystretch}{1.0}
\begin{tabular*}{\columnwidth}{@{\extracolsep{\fill}} p{20mm} p{6mm} p{2mm} p{2mm} p{0.01mm} p{2mm} p{2mm} p{0mm}}
\toprule
&\textbf{Norm.}& \multicolumn{3}{c}{\textbf{Sum of Squared }} & \multicolumn{3}{c}{\textbf{Normalized }} \\
&\textbf{Cost}& \multicolumn{3}{c}{\textbf{ Error [m$^2$]}} & \multicolumn{3}{c}{\textbf{Cost $\times$ Error}} \\
& &\,\,\,\,Low & \,\,\,\,Med. & \,\,\,\,High & Low & Med. & High \\ 
\midrule
Random  & 
 \,0.20 &
 \,\,\,\,4.50 &
 \,\,\,\,0.88 &
 \,\,\,\,0.72 &
 0.90 &
 0.18 &
 0.14 \\[1mm]
Min. Max. Accel. &
 \,\textbf{0.02} &
\,\,\,\,4.79&
\,\,\,\,1.54 & 
\,\,\,\,1.17& 
 \textbf{0.10} &
 \textbf{0.03} &
 \textbf{0.02} \\[1mm]
Deterministic \cite{preiss_simultaneous_2018} &
 \,1.00 &
 \,\,\,\,4.32 &
 \,\,\,\,\textbf{0.85} &
 \,\,\,\,\textbf{0.59} &
 4.32 &
 0.85 &
 0.59 \\[1mm]
Stochastic \cite{frey_online_2020} &
 \,0.10 &
 \,\,\,\,\textbf{4.27} &
 \,\,\,\,0.96 &
 \,\,\,\,0.70 &
 {0.43} &
 {0.10} &
 0.07 \\
\bottomrule
\end{tabular*}
\end{table}

\subsection{Discussion} 
\label{subsec:discussion}

Overall, our results demonstrate that observability-based metrics show promise for trajectory optimization.
We found that, in general, higher-acceleration trajectories produce greater system excitation leading to increased self-calibration accuracy. 
At the same time, minimizing acceleration is desirable for efficiency and safety reasons.
Therefore, trajectories that improve self-calibration accuracy with a reduced overall acceleration profile are preferred.
 
Trajectories optimized using either of the observability-aware methods outperformed random trajectories when both cost and calibration performance were considered. 
Our experiments indicated that the deterministic algorithm of \cite{preiss_simultaneous_2018} produced the highest-accuracy extrinsic calibration.
However, we found that this method also tended to produce higher-cost trajectories.
In contrast, the stochastic algorithm of \cite{frey_online_2020} produced marginally inferior calibration accuracy while significantly reducing the trajectory cost. 
We posit that this may be because the method in \cite{frey_online_2020} takes uncertainty into account, and uncertainty increases with increasing trajectory cost
(i.e., for aggressive flight).

The sensitivity of the optimization algorithms to variations in the `quality' of exteroceptive measurements is also important for real-world applications.
We observed that a substantial reduction in the number of exteroceptive measurements was necessary to significantly affect performance.
On average, the stochastic metric had the largest drop in performance as exteroceptive measurement quality was degraded. 
This result, however, might be explained in part by the stochastic metric producing lower-cost trajectories.
The authors acknowledge that variations in the exact configuration of the simulations, and the constraints of each algorithm, make it difficult to draw overarching conclusions for all trajectory optimization problems.

\section{Conclusion}
\label{sec:conclusion}

We have presented a comparison of two state-of-the-art, observability-aware trajectory optimization methods, and reviewed some of the limitations in the use of observability as an optimization criterion.
To evaluate the optimization algorithms, we leveraged a detailed physics simulation environment where the number and quality of exteroceptive measurements was controlled to be consistent across trials.
This approach permitted an analysis of the effects of variations in the fidelity and accuracy of exteroceptive perception. 
Our results demonstrate that the observability-aware metrics do enable better use of system energy to maximize the knowledge gained about a subset of the system parameters.
For the case of self-calibration, we showed that a minimum level of `perception quality' should be maintained at all times and that trajectories with greater cost are required to achieve the same level of calibration accuracy as overall perception quality degrades.

There are many opportunities for future work.
Numerous variations of our analysis are possible, including adjusting the order of the Lie derivatives considered, the exteroceptive sensor employed, the limits on the system dynamics, the trajectory parameterization, and others.
Additionally, it would be beneficial to study the impact of the estimator used, to determine if different estimation strategies are able to capture more of the benefits that the observability-aware trajectory metrics provide.
More broadly, calibration algorithms, at their core, are reliant on the degree of perceptual uncertainty.
Therefore, future work to better predict measurement uncertainty and quality will likely be valuable for observability-aware trajectory optimization.

\bibliographystyle{IEEEcaps}
\bibliography{refs}

\end{document}